\documentclass[11pt,a4paper]{article}
\usepackage[hyperref]{acl2021}[accepted]
\usepackage{times}
\usepackage{latexsym}
\usepackage{times}
\usepackage{latexsym}
\usepackage{graphicx}
\usepackage{amsmath}
\usepackage{tabularx}
\usepackage{appendix}
\usepackage[utf8]{inputenc} 
\usepackage[T1]{fontenc}    
\usepackage{hyperref}       
\usepackage{url}            
\usepackage{booktabs}       
\usepackage{amsfonts}       
\usepackage{nicefrac}       
\usepackage{microtype}      
\usepackage{algorithm} 
\usepackage[algo2e]{algorithm2e}
\usepackage{natbib}
\usepackage{wrapfig}
\usepackage{pbox}
\usepackage{algpseudocode}
\usepackage{enumitem}

\usepackage{url}
\usepackage{tabu}
\usepackage{amsthm}
\usepackage{float}
\usepackage{stfloats}

\usepackage{microtype}
\SetKwProg{Init}{init}{}{}
\aclfinalcopy 

\title{PoBRL: Optimizing Multi-Document Summarization by \\ Blending Reinforcement Learning Policies}

\author{Andy Su \\ Princeton University \\  \And Difei Su \\ University of British Columbia
        \AND
        John M. Mulvey \\ Princeton University \\  \And
        H. Vincent Poor \\ Princeton University}
\date{}
\begin{document}
\maketitle
\begin{abstract}
 We propose a novel reinforcement learning based framework $\textbf{PoBRL}$ for solving multi-document summarization. PoBRL jointly optimizes over the following three objectives necessary for a high-quality summary: importance, relevance, and length. Our strategy decouples this multi-objective optimization into different sub-problems that can be solved individually by reinforcement learning. Utilizing PoBRL, we then blend each learned policies together to produce a summary that's a concise and complete representation of the original input. Our empirical analysis shows state-of-the-art performance on several multi-document datasets. Human evaluation also shows that our method produces high-quality output.
\end{abstract}
\section{Introduction}
\textbf{Summarization} is the process of creating a concise and comprehensive representation from a large and complex information input. It has important applications in information retrieval (IR) and natural language processing (NLP) domains. Traditionally, summarization can be broadly categorized as extractive or abstractive summarization ${}$\cite{text-summarization}.  In extractive text summarization, salient information and key sentences are extracted directly from the original text without modification. In abstractive text summarization, summary is built by paraphrasing sentences or generating new words that are not in the original text.

In this paper, we tackle extractive summarization within a cluster of related texts (i.e., multi-document summarization\footnote{Note: multi-document here means multiple documents about the same topic.}). Unlike single-document summarization, redundancy is particularly important because sentences across related documents might convey overlapping information. Thus, sentence extraction in such setting is difficult because one will need to determine which piece of information is relevant while avoiding unnecessary repetitiveness. In this sense, finding an optimal summary can be viewed as a combinatorial optimization problem. One way to solve this is the \emph{Maximal Marginal Relevance} (MMR) algorithm \cite{mmr}. In the MMR approach, each sentence is extracted by maximizing its relevance while minimizing over redundancy. Since MMR is an iterative and greedy algorithm, it only achieves local optimality. As another approach, in \cite{global-mmr} the author developed a globally optimal MMR method by reformulating it as an inference problem and solving it using integer linear programming.

In this paper, we present a novel approach to tackle multi-document summarization. We formulate this as a multi-objective optimization problem and we seek to jointly optimize over the following three key metrics necessary for a high quality output  \cite{global-mmr}:
\begin{itemize}[noitemsep]
\item Importance: only the relevant and critical information should be extracted;
\item Redundancy: the extracted information should not be self-repetitive;
\item Length: the final output should be as concise as possible.
\end{itemize}

Our approach optimizes these three objectives simultaneously and is able to produce output which is a condensed and complete summary of the original content. At a high level, our method utilizes MMR to navigate through a complicated and overlapping multi-document space. We present our \textbf{PoBRL} (\textbf{Po}licy \textbf{B}lending with maximal marginal relevance and \textbf{R}einforcement \textbf{L}earning) framework to decouple this multi-objective optimization problem into smaller problems that can be solved using reinforcement learning. Then through our PoBRL framework, we present a policy blending method that integrates the learned policies together so that the combined policy is of high quality in the context of the three identified objectives. Through our newly proposed model and empirical evaluation on the Multi-News \cite{Multinews} and DUC-04 datasets, our approach demonstrates state-of-the-art results. Our \textbf{contributions} are as follows:
\begin{itemize} [noitemsep]
  \item We propose a novel PoBRL algorithmic framework that jointly optimize over the three identified metrics by objective decoupling, independent policy learning. We combine each learned policies by "blending"; 
  \item We propose a novel, simple,  and effective multi-document summarization model that leverage the hierarchical structure of the multi-document input;
  \item Our empirical performance on multi-document datasets shows state-of-the-art performance. Human evaluation also shows that our method produces fluent and high-quality output.
\end{itemize}

\section{Related Work}
Extractive multi-document summarization can be viewed as a discrete optimization problem. In this setting, each source sentence can be regarded as binary variables. Under a pre-defined summary length as the constraint, the optimization problem needs to decide which sentences to be included in the final summary. Several techniques were introduced to solve this problem. For instance, \cite{lin-bilmes-2010-multi} proposed to solve with maximizing sub-modular functions under budget constraint. \cite{global-mmr}\cite{gillick-favre-2009-scalable} reformulated the problem and solved with integer linear programming.

With the recent advances in machine learning, deep learning techniques have gained traction in summarization. For instance, a more recent approach combines BERT with text-matching for extractive summarization \cite{zhong2020extractive}. In \cite{lebanoff-etal-2018-adapting}\cite{zhang-etal-2018-adapting}, the authors adapted the network trained on single document corpus directly to multi-document setting. In \cite{MGS}, the author proposed multi-granularity network for multi-document summarization. In \cite{yasunaga2017graphbased}, this problem was solved by graph convolutional network. Other related works include \cite{oceanv} \cite{cao2016improving}\cite{isonuma-etal-2017-extractive}\cite{nallapati2016summarunner}.

Alternatively, there is RL approach to solve summarization. In \cite {Narayan2018}, the author utilized the REINFORCE \cite{REINFORCE} (an RL algorithm) to train the model for extractive single-doc summarization. A similar approach was also taken in \cite{arumae2018reinforced} by including a question-focus reward. In \cite{FastRL}, the author performed abstractive summarization following the sentence extraction. In \cite{paulus2017deep}, REINFORCE is being applied for abstractive summarization. A more stabilize strategy (Actor-critic \cite{A2C}) has been considered in \cite{li2018actorcritic}. Similar approach has also been investigated in \cite{dong2019banditsum} \cite{pasunuru2018multireward}. However, all these approaches focus on training a \emph{single} RL policy for solving the summarization problem.


To the best of our knowledge, our PoBRL approach that blends \emph{two} RL policies (with one policy optimizes for importance, and another policy optimizes for redundancy) for solving the multi-document summarization problem has not been explored in any of the NLP literature. The closest possible related work is \cite{mix-and-match}, in which the authors proposed a \emph{curriculum learning based} RL method for compressing (mixing) multiples RL policies to solve a high-dimensional (\textbf{non-NLP}) task. Different than their curriculum learning approach, here we show that we can blend(mix) different policies by simply taking the MMR combination of them as in eqn.\ref{eqn:mmr} (and eqn.\ref{eqn:pobrl}), a much simpler and intuitive approach for solving a \emph{NLP} task (multi-doc summarization) without introducing an extra layer of complexity.

\section{Background}
\subsection{Maximal Marginal Relevance}
We leverage MMR to navigate the complicated and overlapping sentence space in the multi-document. MMR provides a balance between relevance and diversity.  Formally, MMR is defined as:

\begin{equation}
\begin{split}
   \text{MMR} =   \text{max}_{s_i \in R\backslash S}( \lambda     \textit{Importance}(s_i) \\
    - (1-\lambda) \text{max}_{s_j \in S} \text{ } \textit{Redundancy}(s_i,s_j) ) 
\end{split}
\label{eqn:mmr}
\end{equation}

where $S$ is the produced summary, $s_i$ is the $i^{th}$ sentence from the documents $R$ excluding $S$, and $\lambda$ is a parameter balancing between importance and diversity. In general, importance and redundancy functions can be in any form. In practice, eqn \ref{eqn:mmr} is implemented as an iterative and greedy algorithm that loops through each sentence in the documents for a pre-defined summary length. The algorithm is shown in Appendix Section A. 

\subsection{Motivation for Reinforcement Learning}
As noted by  \cite{global-mmr}\cite{lin-bilmes-2010-multi}, such iterative and greedy approach is non-optimal because the decision made at each time step is local and doesn't consider its downstream consequences. The sequential nature of algorithm naturally fits within the reinforcement learning framework in which each sentence extraction action is optimized for future(downstream) cumulative expected reward.

  
    \begin{figure*}[b]
    \center{\includegraphics[width=0.89\textwidth]
    {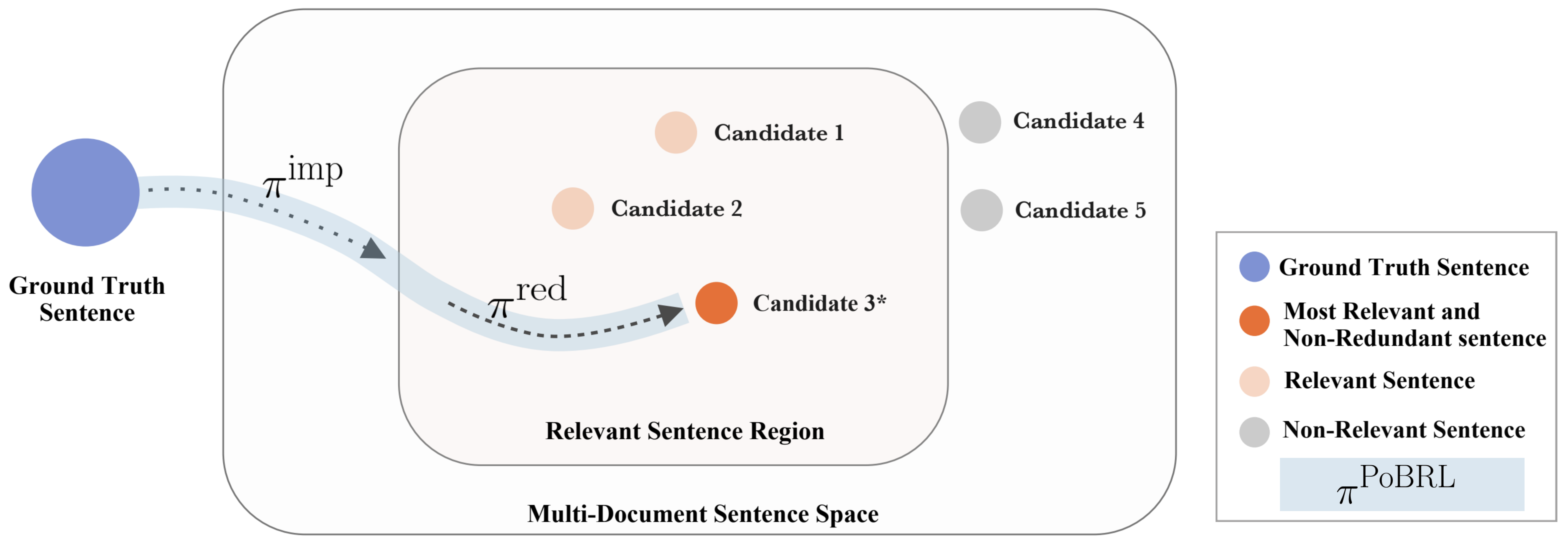}
    }
    \caption{\label{fig:search-multi} Multi-document search space shows sentences across multiple related texts can represent the same ground truth sentence.  The search procedure for $\pi^{\text{PoBRL}}$ is to have $\pi^{\text{imp}}$ get to the relevant sentence region. Then from here, $\pi^{\text{red}}$ finds the most optimal candidate sentence. }
  \end{figure*}

Formally, we view the sentence extraction as a Markov Decision Process (MDP). We define a state $X_t = \{R, S\}$, which is a combination of input documents and extracted set of sentences at time $t$ respectively. At each time step, our policy makes a decision $a_t$ determining which sentence $s_t$ is to be extracted. We define our policy $\pi$ that \emph{searches} over possible sentences across the space. Instead of having a pre-determined length of summary, we let our policy $\pi$ to optimize it directly. We define actions of two types:
\[
    a_t= 
\begin{cases}
    s_t \sim P_\pi(. | X_t ),& \text{if continue to extract} \\
    \text{STOP extraction},              & \text{otherwise}
\end{cases}
\]


With this action setup, our extractive agent can design the optimal strategy as when to stop extraction. When it instead decides to continue extracting sentences, each sentence in the input will be extracted according to the probability $P_{\pi}(\cdot | X_t)$, where $X_t$ is the state which is used to keep track of what has been extracted so far ($S$), and what has not be extracted ($R$). Our goal then is to come up with an optimal extraction policy $\pi^* = \text{argmax}_{\pi \in \Pi} E(\sum_t r_t | X_t, a_t)$, where $r_t$ is the reward contribution of adding sentence $s_t$ onto the system summary $S$.

If we define the reward contribution to be the ROUGE score \cite{lin-2004-rouge} that we wish to optimize, then it can be shown (see Appendix \ref{section:decouple}) that :\begin{equation}
    \text{ROUGE}(S) \approx \sum_{i=1} \text{imp}(s_i) - \sum_{a,b \in S, a \neq b} {\text{red}(a,b)}
    \label{eqn:decouple}
\end{equation}
Fig.\ref{fig:search-multi} shows the sentence search process for the multi-document space. For each ground truth sentence (labeled as blue), there might exist more than one candidate sentences coming across multiple related text. For the optimal summary $S^*$, we want to extract the single best sentence (colored dark orange) from a pool of similar looking sentences (region denoted as"relevant sentence space"in the figure). 

For effective navigation in such a search space, we proposed our PoBRL framework(Algorithm \ref{alg:the_alg}).  Based on eqn \ref{eqn:decouple}, our framework decomposes the importance and redundancy objectives independently into two sub-problems. Since each problem has its own local objectives, to maximize eqn \ref{eqn:decouple}, we simply design policies $\pi ^{\text{imp}}$ and $\pi^\text{red}$ to optimize for each objective independently. At a high level, we can think of $\pi^\text{imp}$ helps the policy to narrow down the search by identifying the most relevant(or important) regions. From these regions, $\pi ^{\text{red}}$ searches for the most relevant and diversify (non-redundant) sentence.  This search procedure is graphically illustrated in Fig. \ref{fig:search-multi} where $\pi^\text{imp}$ leads the search toward the region containing candidates (1,2,3) which are sentences relevant to the ground-truth sentence; then in this region, $\pi^\text{red}$ leads the search to discover the most relevant and non-redundant sentence (denoted as candidate 3*). Connecting $\pi^\text{red}$ and $\pi^\text{imp}$ together formulates the optimal policy $\pi^\text{PoBRL}$ which selects the most optimal sentence.

\section{PoBRL Algorithmic Framework}
We present our PoBRL algorithmic framework as in Algorithm 1. 
We start off the algorithm by independently training two policies $\pi^{\text{imp}}_{\theta}$ and $\pi^{\text{red}}_{\phi}$, where $\theta$ and $\phi$ represent two sets of neural network parameters. We trained both policies with actor-critic (Section.\ref{sec:a2c}) but with different objective. For the $\pi^{\text{imp}}_{\theta}$, it is trained by optimizing over \emph{importance}. For the $\pi^{\text{red}}_{\phi}$, it is trained with \emph{redundancy} as its objective. 


    

\SetKwInput{KwInput}{Input} 

\begin{algorithm}[H]
\small
\SetAlgoLined
\KwInput{$\lambda, R$}
{
\small
 S$\leftarrow \{\}$ \\
 $\pi^\text{imp}_{\theta} \leftarrow $train-RL-Policy(importance) as in Section \ref{sec:imp}\\
 $\pi ^ \text{red}_{\phi} \leftarrow $ train-RL-Policy(redundancy) as in Section \ref{sec:red} }\\ \\
 \While{$\text{True}$}{
  $X_t = \{R, S\} $\\
  $P^{\text{imp}}_{1:{nm}} \leftarrow $  $P_{\pi^{\text{imp}}_{\theta}}(\cdot |X_t)  $ with $\pi^{\text{imp}}_{\theta}$ (as in Eqn.\ref{eqn:sentence_probability}) \\
  $P^{\text{red}}_{1:{nm}} \leftarrow $  $P_{\pi^{\text{red}}_{\phi}}(\cdot |X_t)  $ with $\pi^{\text{red}}_{\phi}$ (as in Eqn.\ref{eqn:sentence_probability}) \\
  $\text{(optional) } \text{calculate a new value of }\lambda$ (Eqn.\ref{eqn:advantage_decode}) \\ 
  {\small
  calculate $\pi^\text{PoBRL}$ by blending $\pi^{\text{imp}}_{\theta}$, $\pi ^ \text{red}_{\phi}$ (Eqn.\ref{eqn:pobrl})
  }\\
  $P^{\text{PoBRL}}_{1:{nm}} $$\leftarrow P_{\pi^{\text{PoBRL}}}(\cdot | X_t)$ with $\pi^{\text{PoBRL}}$as in Eqn.\ref{eqn:mmr_probability} \\
  $s_{\text{select}} \sim  P^{\text{PoBRL}}_{1:{nm}}$ (extract sentence by sampling from extraction probability)
 $S \leftarrow S \cup  \{s_{\text{select}} \}$ \\ \;
 \If{ $\text{STOP}$ $\sim $ $\pi^{\text{PoBRL}} (. |X_t)$}{
    \text{break}\
 }
 }
 \caption{PoBRL Algorithmic Framework}
 \label{alg:the_alg}
return system summary $S$
\end{algorithm}
 Assume there are $n$ articles and $m$ sentences per article for a total of $nm$ number of sentences in the input, once we have trained these two policies, we then calculate the sentence extraction probabilities for each sentence in the input as: 
 \begin{equation}
    P_{1:{nm}} \leftarrow P_{\pi}(s_k | X_t) \text{ for all } k \in {nm}
  \label{eqn:sentence_probability}
 \end{equation} where $P_{1:{nm}} \in R^{nm \times 1}$ (with each entry represents the sentence $s_k$ extraction probability). We first calculate this sentence extraction probability from eqn.\ref{eqn:sentence_probability}  with $\pi^{\text{imp}}_{\theta}$ as:  $P^{\text{imp}}_{1:{nm}} \leftarrow $  $P_{\pi^{\text{imp}}_{\theta}}(\cdot |X_t)$. We repeat this calculation with $\pi^{\text{red}}_{\phi}$ as: 
$P^{\text{red}}_{1:{nm}} \leftarrow P_{\pi^{\text{red}}_{\phi}}(\cdot |X_t)$.
 
Now, we are ready to blend these two policies together utilizing MMR, which is:
\begin{equation}
\small
    \pi^\text{PoBRL}(\cdot | X_t ) = \lambda \pi^{\text{imp}}_{\theta}(\cdot | X_t) - (1-\lambda) \pi^{\text{red}}_{\phi}(a_{2,t} | X_t,P^{\text{imp}}_{1:{nm}})
\label{eqn:pobrl}
\end{equation}
where the sentences extraction probabilities of this newly blended policy PoBRL are given by:
\begin{equation}
 P^{\text{PoBRL}}_{1:{nm}} \leftarrow P_{\pi^{\text{PoBRL}}}(\cdot | X_t) 
 = \lambda P^{\text{imp}}_{1:{nm}} - (1-\lambda) P^{\text{red}}_{1:{nm}}   
\label{eqn:mmr_probability}
\end{equation}

Now comparing both Eqn.\ref{eqn:mmr_probability} and Eqn.\ref{eqn:pobrl}  to the original MMR setup as in Eqn.\ref{eqn:mmr}, we see that they are nearly identical in terms of setup. Finally, to determine which sentence to extract at the $t=1$ (the first time step), we sample $s_{\text{select}} \sim  \pi^{\text{PoBRL}} ( \cdot | X_{t=1})$, with the sentence extraction probabilities given by Eqn.\ref{eqn:mmr_probability}. We appended the extracted sentence $s_{\text{select}}$ into the summary set $S$, and we update the state $X_{t=1} =\{R, S\}$. We repeat the extraction process until stop action is sampled.

\subsection{Reinforcement Learning Training Algorithm: Actor-Critic}
\label{sec:a2c}
To generate labels, we utilize the following technique. For each sentence $gold_k$ in gold-summary, we find its most similar sentence from the input text.  We define similarity using the \text{ROUGE-L} measure. That is, $ \text{   label}_k 
= \text{argmax} \text{  ROUGE-L}(gold_k, s_k)$, for each $s_k$ in the example. We define the rewards in corresponding to the agent's actions. Formally, \begin{equation}
\small
    r_t= 
\begin{cases} \label{eqn:reward}
    \text{ROUGE-L}(s_k, \text{gold}_k),& \text{if extract} \\
    \text{ROUGE-1}(\text{summ}_\text{sys}, \text{summ}_\text{gold}),              & \text{if stop}
\end{cases}
\end{equation}


where $\text{summ}_\text{sys}$ is the system summary and $\text{summ}_\text{gold}$ is the gold summary.
We utilize actor-critic\cite{actor-critic} to train our policy network. The gradient update for the weights of policy network will be:
\begin{equation}
\small
\nabla J(\psi) = E_{\pi_{\psi}} [\nabla_{\psi} \text{log} \pi_{\psi} (a_t|X_t) A(X_t, a_t)]
\label{eqn:a2c}
\end{equation}
where $\pi_{\psi}$($\cdot$) is the policy function that takes an input state and outputs an action (in our case, which sentence to extract), $\psi$ consists of the learnable parameters of the neural network. Next, we have
\begin{equation}
\small
A(X_t, a_t=s_t) = Q(X_t, a_t=s_t) - V(X_t) 
\label{eqn:advantage}
\end{equation}
is the advantage function which measures the \textbf{advantage} of extracting a \emph{particular sentence} $s_t$ over the rest of the sentences in the pool. And $V(X_t) = E_{\pi_{\theta}}\{\sum \gamma^t r_{t+1} |X_t\}$ measures the quality of state,  $Q(X_t, a_t=s_t) = E_{\pi_{\theta}}\{\sum \gamma^t r_{t+1} |X_t, a_t=s_t\}$.

We instantiate two sentence extraction policy networks as in Algorithm \ref{alg:the_alg}, each with its own objective, and trained with actor-critic.

\subsubsection{train-RL-Policy(Importance)}
\label{sec:imp}
We train our first policy network $\pi^{\text{imp}}_{\theta}$ to maximize the \textit{importance} over each step with actor-critic as giving by eqn.\ref{eqn:a2c}. To do this, we modify the first part of the reward eqn.\ref{eqn:reward} to be ROUGE-L$_{F_1}$ to encourage this RL policy network searching for the most relevant sentence region from the multi-document space. 

\subsubsection{train-RL-Policy(Redundancy)}
\label{sec:red}
We train our second policy network $\pi^{\text{red}}_{\phi}$ with the \textit{redundancy} as its objective using actor-critic. To do this, we modify the reward measure of eqn.\ref{eqn:reward} (first part) to be $ \text{ROUGE-L}_{\text{Precision}}$ to encourage this policy network to identify similarity between each sentences in the pool versus the ones already extracted. This in effect helps $\pi^{\text{imp}}_{\theta}$ to identify the best (or the most relevant but non-redundant) sentence from the region identified by $\pi^{\text{imp}}_{\theta}$. 





\subsection{Adaptive Tuning of $\lambda$ with the Advantage Function}

The formulation in eqn.\ref{eqn:pobrl} and eqn.\ref{eqn:mmr_probability} are good as long as we have a reasonably good $\lambda$. In practice, the $\lambda$ parameter can be a fixed numerical value that can be computed offline and determined beforehand. In reality, this is not an easy task because: 1) the $\lambda$ parameter might vary from document to document; and 2) the optimal value of $\lambda$ might change throughout the extraction process. 

To solve this problem, we propose an innovative idea that requires no separate offline training process and can be computed online in an adaptive fashion by leveraging the power of the RL framework. When we train the network with actor critic, we get the advantage function automatically as a byproduct. Therefore, the advantage function from eqn $\ref{eqn:advantage}$ provides a quality measure by comparing sentences from the pool. Utilizing the advantage function, we can have a score function that rates each sentence in accordance with the importance and redundancy/diversity balance. Formally, 
\begin{equation}
\lambda_t^{\text{adv}} = A^{\pi^\text{imp}_{\theta}} (X_t, a_t=s_t) - A^{\pi^\text{red}_{\phi}} (X_t, a_t=s_t) 
\label{eqn:advantage_decode}
\end{equation}

\noindent where $A^{\pi^\text{imp}_{\theta}}, A^{\pi^\text{red}_{\phi}}$ are the advantage functions of the importance and redundancy policy network respectively, and $a_t$ defines the action of extracting a particular sentence at time $t$.  The advantage function defined in eqn.$\ref{eqn:advantage}$ scores sentences from the pool according to the $imp$ or $red$ measure. Note that since the sentences already extracted are encoded by the state $X_t$, so the advantage function is context aware.


\section{Hierarchical Sentence Extractor}

  \begin{figure} [htbpp]
    \center{\includegraphics[width=1.05\columnwidth]
    {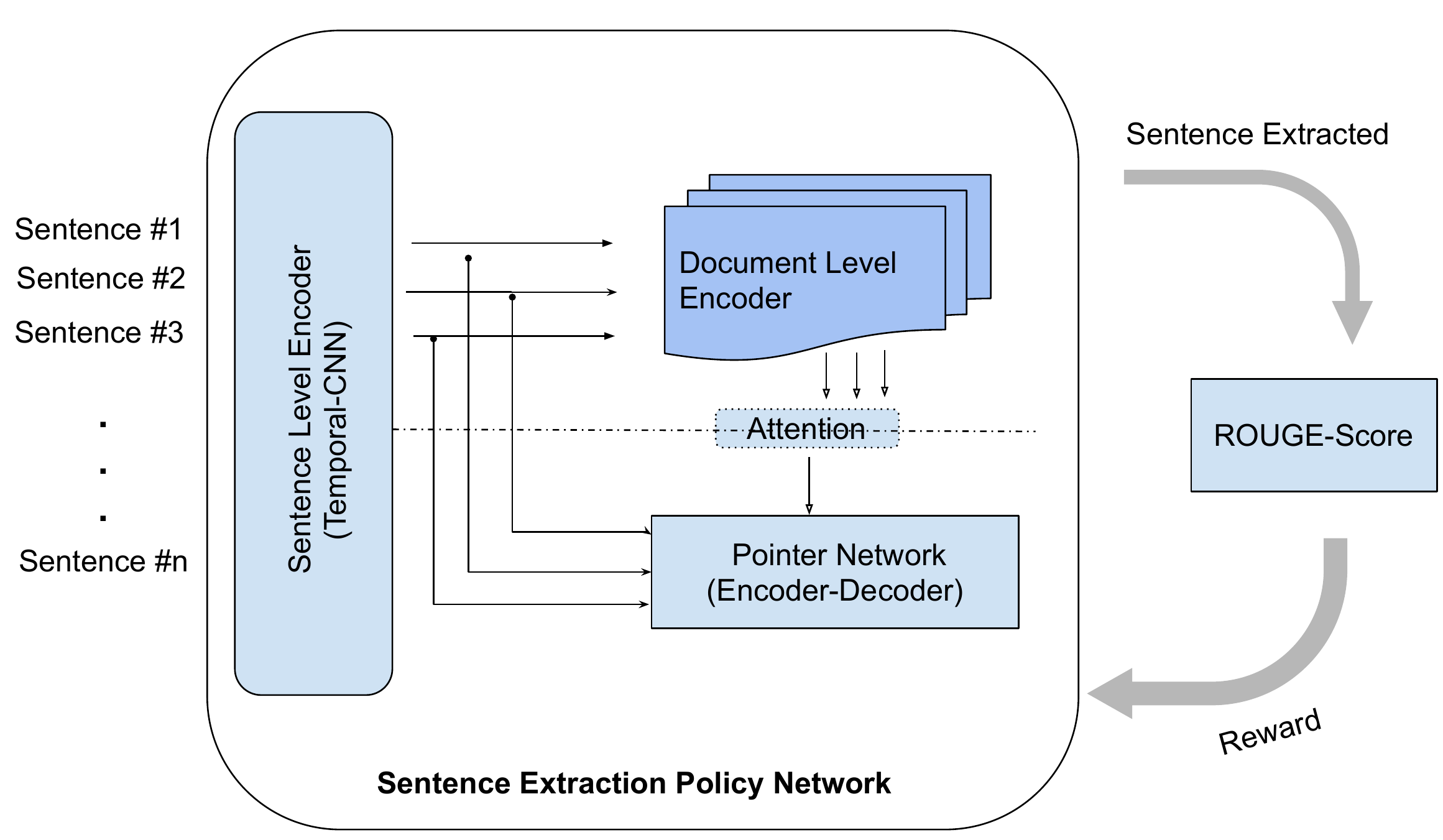}}
    \caption{The Hierarchical RL Sentence Extractor Model. Each multi-document input will go through the article level path (dark blue) and the sentence level path (light blue).}
    \label{fig:policy-model}
  \end{figure}
In this section, we describe our model structure. Our hierarchical sentence extractor consists of the following building blocks: sentence encoder, document-level encoder, attention, and pointer network. The complete model structure is shown in Fig \ref{fig:policy-model}. At the high level, the model operates like a pointer network with the attention module adjusts the focus between the article level and sentence level. 

\subsection{Sentence Encoder}
Our model deals with each input document by reading one sentence at a time. Each input sentence will first be encoded by the sentence encoder. We use a temporal-CNN to encode each sentence, and denoted each sentence encoded hidden state with $s^{art=j}_{enc_i}$, where $art=j$ indicates the sentence that came from the $j_{th}$ article and $i$ represents the sentence number from that article. Each of the encoded sentences will then go through two paths: the article-level path and the sentence aggregation path.

\subsection{Article-Level Path}
In our multi-document setting, each input contains one or more articles. Here, we are looking to formulate a representation for each article of the multi-document input. The article level path represents the dark blue colored region in Fig \ref{fig:policy-model}. We implement the article-level encoder with a bidirectional LSTM. For each article of the multi-document input, this encoder takes its input sentences (eg,  $s^{art=1}_{enc_1}$, $s^{art=1}_{enc_2}, .... s^{art=1}_{enc_n}$). This article is represented by the corresponding article level sentence encoded hidden states:
$h^{art=j}_i = \text{LSTM}(s^{art=j}_{enc_i}, h^{art=j}_{i-1})$
for $i \in [1, n]$. We repeat this process for all the article of the multi-document input text.

\subsection{Sentence Aggregate Path}

In this path, we feed in each sentence $s^{art=j}_{enc_i}$ for $i\in [1, n], j \in [1, m]$ one by one by concatenating them together as if they came from a single piece of text. We utilize a pointer network \cite{pointernn} that captures local contextual and semantic information. This pointer network utilizes information by the article level encoder and form the basis for sentence extraction.
We implement the pointer network with encoder-decoder based LSTM. Formally, we can write:
\begin{equation} \label{eq2}
u^t _k = v^T \text{tanh}(W_{sent} e_k + W_{art} c_t)
\end{equation}
where $c_t$ is the output of the attention mechanism at each output time $t$, $e_k$ is the hidden state of the encoder pointer network LSTM. In other words, ($e_1$, $e_2$, $e_3$, ..., $e_k$, ..., $e_{n \times m}$) are the encoder hidden states for the input sentences $s^{art=j}_{enc_i}$ for $i\in [1, n], j \in [1, m]$. Lastly, $v, W_{sent}$, and $W_{art}$ are the network learn-able weight matrices.

\subsection{Attention}
The attention module integrates article level information to the sentence aggregate level extraction network.We implement this module with dot product attention \cite{luong2015effective}. 

Let $d_t$ to be hidden state of decoder pointer network LSTM at output time index $t$, $h_k$ to be the corresponding article level sentence encoded hidden state, then: 
\begin{equation}
\small
  c_t = \sum_k \alpha_{t,k} h_k,  
\qquad
\alpha_{t,k} = \frac{e^{(\text{score}(d_{t}, h_k))}}{\sum_{k'} e^{(\text{score}(d_{t}, h_{k'}))}},
\qquad
\label{exv:eqn:UmrechnungEingangsgroesse3}
\end{equation}

and that $\text{score}(d_t, h_k) = d_t^Th_k $.

\subsection{Actor Critic Sentence Extraction Policy Network}
Now, we can connect the dots between each component above, and this builds our sentence extractor. During the actual sentence extraction process, we ranked each candidates in the pool set by:
$
P(s_k | s_{k-1}, ... s_0) = \text{softmax}(u_k), \text{for } k \in \text{pool set}.$ The pool set is initiated to contain all of the sentences in the input text. When a particular sentence has been extracted, we remove it from the pool set. That is: $\text{Pool Set } \leftarrow \text{Pool Set } \setminus \{s_k\}$
\subsection{Training Details}
Training a reinforcement learning neural network from scratch is difficult and time consuming because the initial policy network tends to behave randomly. To accelerate this learning process, we warm start it with supervised learning. 


\textbf{Warm Start: Supervised Learning}
We modify the training objective to be negative log-likelihood, and we train our network by maximizing this objective function. 
After pre-trained supervised learning, we can then train with actor-critics.

\section{Experimentation}
We showcase the performance of our methods on the following two multi-document datasets: Multi-News and DUC 2004. We begin by defining the following baselines :

\noindent \textbf{Lead-n}. We concatenate the first $n$ sentences from each article to produce a system summary. 


\noindent \textbf{MGSum (ext/abs)}, an extractive/abstractive summarization method with Multi-Granularity Interaction Network. \cite{MGS}.

\noindent \textbf{MatchSum}, an extractive summarization method with text-matching and BERT \cite{zhong2020extractive}.

\noindent \textbf{GRU+GCN}: a model that uses a graph convolution
network combined with a recurrent neural network
to learn sentence saliency \cite{yasunaga2017graphbased}.



\noindent \textbf{OCCAMS-V}  a topic modeling method \cite{oceanv}.

\noindent \textbf{CopyTransformer} This is a transformer model  from \cite{gehrmann-etal-2018-bottom} that implements the PG network.

\noindent \textbf{Hi-Map} This is a hierarchical LSTM models with MMR attention \cite{Multinews}.

\noindent \textbf{FastRL-Ext} This is the reinforcement learning approach with extractive summarization focusing on single document summarization \cite{FastRL}. Here, we adopt it and apply it to the multi-document case.

\noindent \textbf{RL w/o Blend} We train the multi-doc Hierarchical Sentence Extractor (this paper) with actor-critic \emph{without} policy blending (i.e., single policy instead of blending of two policies). This can be regarded as setting  $\lambda=1.0$ in PoBRL method (eqn.\ref{eqn:mmr_probability} and eqn.\ref{eqn:pobrl}) which optimize purely on importance. 

\subsection{Experimental Setup}
We report the ROUGE $F_1$ score on with ROUGE-1, ROUGE-2, ROUGE-SU(skip bigrams with a maximum distance of four). For our PoBRL model, we instantiated two actor-critic policy networks representing $\pi^{\text{imp}}_{\theta}, \pi^{\text{red}}_{\phi}$ respectively. We then blend these two polcies by forming $\pi^{\text{PoBRL}}$ as in eqn.\ref{eqn:pobrl} with extraction probability in eqn.\ref{eqn:mmr_probability}. For complete network hyper-parameter and training details, please refer to Appendix D.

Besides using a fixed $\lambda$ value, we also evaluated a $\lambda$ value calculated using the advantage function as in eqn.\ref{eqn:advantage_decode}, and denote it by PoBRL($\lambda=\lambda_t^\text{adv}$). At each extraction step, the optimal $\lambda$ is recalculated by balancing between the importance and redundancy consideration. 

\begin{table}[htbp]
\begin{center}
\scalebox{0.75}{
\begin{tabular}{l|c c c c}
\toprule
Method                             & R-1 & R-2  & R-L\\ 
\midrule
Lead-3                             & 39.41   & 11.77     & 18.03         \\
CopyTransformer                    & 43.57   & 14.03      & 20.50            \\
Hi-Map                             & 43.47   & 14.87     & 21.38        \\
FastRL-Ext                           & 43.56  & 16.05     & 39.69   \\
MGSum-ext                            & 44.75   & 15.75      & 40.84         \\
MGSum-abs                            & 46.00   & 16.81      & 41.36         \\
MatchSum                            & 46.20   & 16.51       & 41.89           \\

{RL w/o Blend }                                & 44.01      & 17.63  &  33.96     \\
$\textbf{PoBRL}  (\lambda = 0.8) (\textbf{Ours}) $  & {45.13}   & {16.69}      & {41.12}     \\
$\textbf{PoBRL}  (\lambda = \lambda_t^\text{adv}) (\textbf{Ours}) $  & \bf{46.51}   & \bf{17.33}    & \textbf{42.42}     \\

\bottomrule     
\end{tabular}}
\end{center}
\caption{MultiNews dataset}
\label{table:multi-news}
\end{table}

\subsection{Empirical Performance}
We empirically evaluate the performance of the models on the following two multi-document datasets. 

\textbf{Multi-News}
The Multi-News \cite{Multinews}  is a large dataset that contains news articles scrapped over 1500 sites. The average number of words per doc and ground-truth summaries are much  longer than DUC-04. This dataset has 44972 number of training examples, 5622 for validation and test set respectively. We trained our models on this dataset, and reported their performance in Table \ref{table:multi-news}. Our best model PoBRL ($\lambda$ = $\lambda_t^{\text{adv}}$) achieves a score of 46.51/17.33/42.42, outperforms other baselines in all metrics (with the strongest ones being MatchSum and MGSum).  Here, we also show the performance of our model on different values of $\lambda$. When $\lambda = 1.0$, the model reduces to single policy and is not policy-blended (denoted by RL w/o Blend). When $\lambda$ = $\lambda_t ^ \text{adv}$, the value of $\lambda$ is adaptively determined by the advantage function as in eqn. \ref{eqn:advantage_decode}.  For a fixed value of $\lambda=0.8$, we achieve a good score (45.13/16.69/41.12). However, it is intuitive that the optimal value should be varied from document to document, and it might change during the extraction process. That is reflected by the PoBRL($\lambda = \lambda_t^\text{adv}$) advantage method which provides a significant performance boost.  On the other hand, the RL w/o Blend represents a model without policy blend strategy. Comparing RL w/o Blend with PoBRL demonstrates the effect of our policy-blending algorithm. Incorporating policy-blending on top of this hierarchical sentence extraction model provides remarkable performance boost (going from RL w/o Blend's 44.01/15.65/33.96 to PoBRL's 46.51/17.33/42.42).

\textbf{DUC-04} This is a test only dataset that contains 50 clusters with each cluster having 10 documents. For each cluster, the dataset provides 4 human hand-crafted summaries as ground-truth. Following \cite{Multinews}\cite{lebanoff-etal-2018-adapting}, we trained our models using the CNN-DM \cite{CNNDM} dataset, and report the performance of different models on Table \ref{table:DUC}. Our best model (PoBRL $\lambda$ = $\lambda_t^{\text{adv}}$) achieves 38.67/10.23/13.19 (ROUGE-1/2/SU) which is a substantial improvement over all other baselines (with the closest ones being OCCAMS-V and GRU+GCN). On the other hand, using a fixed value of $\lambda=0.8$ of PoBRL still achieves good performance (38.13/9.99/12.85). Comparing PoBRL's $\lambda=0.8$ to $\lambda = \lambda_t^\text{adv}$ shows the power of our method of calculating $\lambda$ with the advantage function. The effect of our policy blending algorithm is demonstrated by comparing RL w/o Blend with PoBRL. Here, we also observe a major performance boost (36.95/9.12/11.66 versus 38.67/10.23/13.19).
\subsection{Human Evaluation}
In this experiment, we randomly sampled 100 summaries of Multi-News dataset from the following models: Hi-Map, CopyTransformer, MatchSum, and PoBRL. Following \cite{chu2018meansum}, we asked judges\footnote{All judges are native English speakers with at least a bachelor’s degree and experience in scientific research. We compensated the judges at an hourly rate of \$28.} (10 for each sample) to rate the quality of system generated system in a numerical rating of 1 to 5, with the lowest score to be worst and the highest score to be the best.  We evaluated the quality of the system generated summarization in terms of the following metrics as in \cite{chu2018meansum} \cite{duc05}: Grammatically, Non-redundancy, Referential clarity, Focus, Structure and Coherence. Our result is shown in Table \ref{table:human}. 

\begin{table*}[htbp]
\begin{center}
\begin{small}
\scalebox{0.75}{
\begin{tabular}{c|c c c c c}
\toprule
Method           &  Grammar  & Non-redundancy & Referential clarity & Focus &  Structure and Coherence \\ \midrule
CopyTransformer    & 3.83  & 3.26   & 3.89    & 3.94  &4.04          \\ 
Hi-Map            &  4.06  & 3.46   & 3.74   &  4.01  & 3.89     \\ 
MatchSum            &  4.26  & 3.32   & 4.12   &  4.00  & 3.96    \\ 
\textbf{PoBRL (Ours)}  &\textbf{4.28} & \textbf{4.68}   & \textbf{4.19}    & \textbf{4.23}  & \textbf{4.08}  \\ \hline
\end{tabular}
}
\end{small}
\end{center}

\caption{Human evaluation on system generated summaries of Multi-News dataset}
\label{table:human}
\end{table*}

\begin{table}[htbp]
\begin{center}
\scalebox{0.85}{
\begin{tabular}{l| c c c}
\toprule
\small
Method              & R-1 & R-2 & R-SU  \\
\midrule
Lead-1               & 30.77   & 8.27    & 7.35              \\ 
CopyTransformer     & 28.54   & 6.38    & 7.22             \\ 
Hi-Map     & 35.78   & 8.90    & 11.43             \\ 
GRU+GCN:             & 38.23   & 8.48    & 11.56            \\
$\text{OCCAMS}_V$             & 38.50   & 9.76    & 12.86             \\ 
RL w/o Blend    & 36.95   & 9.12   & 11.66      \\ 
$\textbf{PoBRL}(\lambda=0.9) (\textbf{Ours})$ & {38.13}   & {9.99}    & {12.85} \\
$\textbf{PoBRL}(\lambda=\lambda_t^{\text{Adv}})( \textbf{Ours})$ & \textbf{38.67}   & \textbf{10.23}    & \textbf{13.19}  \\ 
\bottomrule
\end{tabular}
}
\end{center}
\caption{DUC-04 dataset}
\label{table:DUC}
\end{table}

\begin{table}[htbp]
\begin{center}
\resizebox{\columnwidth}{!}{
\begin{tabular}{c|c c}
\toprule
Redundancy Measures              &   Cosine (Higher the better) &  ROUGE-L (Lower the better) \\
\midrule
RL w/o Blend                  &  12.49  & 2.28 \\
$\text{PoBRL}$ & \textbf{12.73}   & \textbf{0.19}            \\ 
\bottomrule
\end{tabular}
}
\end{center}
\caption{Redundancy Analysis, the first column redundancy is in cosine similarity measure (the higher the better), and the second column is the ROUGE-L measure (the lower the better).}
\label{table:red}
\end{table}

\section{Analysis}
Since the Multi-news dataset provides higher number and more variant/diversified input examples, this will be the dataset which we will be analyzing below with 1000 examples randomly sampled. 
\subsection{Varying number of input document}
\begin{figure}[htbpp]
    \center{\includegraphics[width=0.92\columnwidth]
    {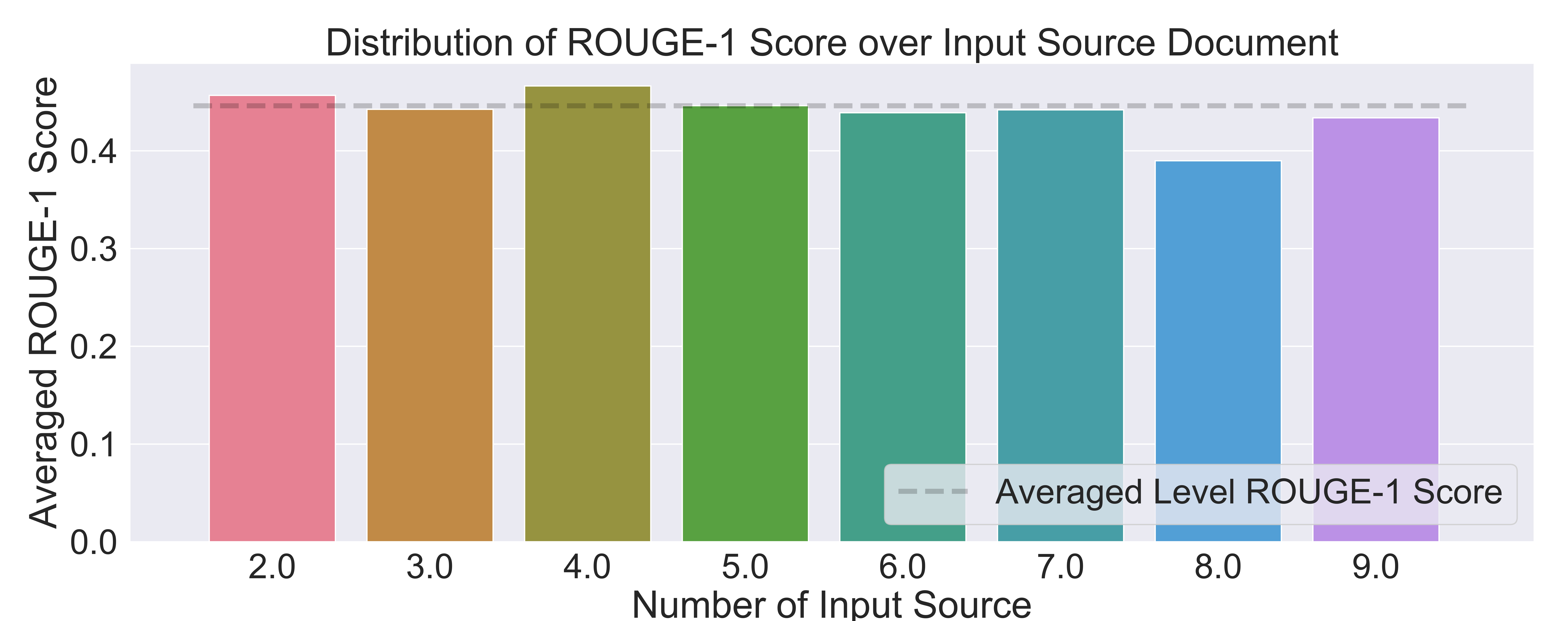}}
    \caption{\label{fig:R1_performance} Model performance on ROUGE-1 on Multi-doc inputs of different number of document.}
  \end{figure}

Unlike DUC04, each example of the multi-news dataset has different number of source documents (range from 2 to 9).  In general, the higher the number of the example, the longer the input. In Fig \ref{fig:R1_performance}, we show the effect of our proposed method in handling different numbers of source document in each multi-doc input. As can be seen in the figure, the performance is almost uniformly distributed, showing that our method can handle higher or lower number of input documents (which also translate to longer or shorter piece of input text) equally well.

\subsection{Redundancy}
We measure the redundancy of our system summary by comparing the policy blending formulation of our model (i.e., PoBRL) against with our non-policy-blend (i.e., RL w/o Blend) model. This comparison allows us to understand the effectiveness of employing the policy-blending strategy. We measure redundancy by making use of the ROUGE-L score and the BERT's \cite{devlin2018bert} embedding, separately. Our evaluation criterion is the following: for each sentence in the generated summary, we compute the redundancy metric between that particular sentence against with the rest of the sentence in the summary. In ROUGE-L experiment, we report average sentence level ROUGE-L score within the input example. PoBRL reports the lowest score of 0.19 versus 2.28 as in RL w/o Blend, signifying the significant reduction in overlap (or redundancy) in the system summary. In the BERT's experiment, we report the average cosine similarity score. As shown in Table \ref{table:red}, the PoBRL model achieves higher cosine measure (signifies more diversify and less repetitive). 



\section{Conclusion}
In this paper, we have proposed a novel $\textbf{PoBRL}$ algorithmic framework that allows independent sub-modular reinforced policy learning and policy blending leverage on maximal marginal relevance and reinforcement learning. We have verified the proposed framework on our proposed model. Our empirical evaluation shows state-of-the-art performance on several multi-document summarization datasets. 

\bibliography{acl2021}
\bibliographystyle{acl_natbib}
\newpage
\newpage
\newpage
\newpage
\clearpage
\appendix
\label{sec:appendix}
\newpage
\newpage
\begin{appendices} 
\section {An Iterative and Greedy MMR Algorithm}
\begin{algorithm}[H]
\SetAlgoLined
\KwResult{System summary}
 S$\leftarrow \{\}$ \\ \;
 \While{\text{len}$ <$ \text{maxLen}}{
  \For{$s_i$ \text{in} $R \backslash S $}{
  \text{score}$_{imp} = $ \text{Importance}$(s_i) $ \\ \;
  \For{$s_j$ \text{in} $S$}{
  \text{score}$_{red}$ = \text{max}(\text{score}$_{red}$, \text{Redundancy}$(s_i, s_j)$ ) \\ \;
  }
  \If{ $\textit{MMR-Score} < \lambda$ \text{score}$_{imp} - (1- \lambda) \text{score}_{red}$
  }{
   $ s_{\text{pick}} = s_i$ \\
   \text{Update MMR-Score}
  }
 }
 $S \leftarrow S \cup  \{s_{\text{pick}}\}$
 }
 \caption{Greedy Iterative MMR Algorithm}
 \label{alg:iterative-mmr}
\end{algorithm}

The MMR from equation \ref{eqn:mmr} is implemented as an iterative and greedy algorithm. The complete procedure is shown in Algorithm \ref{alg:iterative-mmr}.

\section{Optimization Objective Decoupling}
\label{section:decouple}
In this section, we derive the result for decoupling the multi-objective optimization problem into small sub-problems. We borrow the notation and formulation of ROUGE score from \cite{peyrard-eckle-kohler-2016-optimizing}. 

Let $S = \{s_i | i \leq m\}$ be a set of $m$ sentences that constitute a system summary, and let $\rho(S)$ be the ROUGE-N score of $S$, where ROUGE-N evaluates the n-gram overlaps between the gold summary and the system summary $S$. Then, \begin{equation}
    \rho(S) = \frac{1}{R_N} \sum_{g \in S^*} (min(F_s(g), F_s^*(g))
\end{equation}
\noindent where $S^*$ is the gold summary, $F_s(g)$ denotes the number of times that the n-gram of $g$ type occurs over $S$, and $R_N$ denotes the number of n-gram tokens.

Let $C_{Y, S^*} = min (F_Y(g), F_{S^*}(g))$ to be the contribution of the n-gram $g$,  $\epsilon (a \land b) = \sum_{g \in S^*} max(C_{a, S^*} (g) + C_{b, S^*} (g) - F_{s^*} (g), 0) $ to be the redundancy between between $a, b$ in the summary, it can be shown \cite{peyrard-eckle-kohler-2016-optimizing} that we can write 

\resizebox{.9\linewidth}{!}{
  \begin{minipage}{\linewidth}{
\begin{align}
    \rho(S) & =  \sum_{i=1} ^{m} \rho(s_i) \\ 
    & + \sum_{k=2}^{m}(-1)^{k+1}(\sum_{1 \leq i_1 \leq ... \leq i_k \leq m}  \epsilon^{(k)} (s_i \land .. \land s_{i_k})) \\
    & \approx  \sum_{i}^m \rho(s_i) - \sum_{a,b \in S, a \neq b} \Tilde{\epsilon}(a \land b) \\
    &  \approx \sum_{i}^m \text{imp}(s_i) - \sum_{a,b \in S, a \neq b} \text{red}(a,b)  
\end{align}
}
\end{minipage}
}

where $\Tilde{\epsilon} ( \cdot )$ approximates the redundancy, $\text{imp}( \cdot )$ denotes importance function, and $\text{red}( \cdot )$ denotes redundancy function. In other word, to maximize the ROUGE score, we can optimize it by maximizing the importance of each sentence while subtracting(minimizing) over the redundancy. 


Now, we define a policy $\pi$ with the goal of \emph{searching} over the sentence space and picking the most optimal set of sentences $\hat{S^*}$ such that $\pi^* = \text{argmax}_{\pi}  E(\sum_{t}r_t |s_t, a_t)$, where $a_t$ corresponds to searching over the sentence space to determine which sentence to be added onto the summary $S$, $r_t = \rho (s_i)$  is interpreted as the importance (or the ROUGE-score contribution) of adding sentence $s_i$  onto the summary set $S$, and the policy $\pi(a_t|X_t)$ outputs the probability of selecting a particular sentence and adding it onto $S$.  Now, instead of having a pre-defined length $m$, our policy $\pi$ also decides the optimal summary length by determining when to stop the selection extraction.

In the multi-document setting, however, related documents might contain large amount of overlapping sentences.  As shown in Fig.\ref{fig:search-multi}, for each sentence in the ground-truth summary, there exist multiple candidate sentence across related text. The optimal sentence selection strategy is to have two policies, with the first search policy $\pi^{\text{imp}}$  learns to discover for the most relevant sentence space. And from this smaller space, we have the second policy $\pi^{\text{red}}$ learns to search for the most relevant but non redundant sentence (denotes as the most optimal sentence). 

At a high level, these two policies, with $\pi^{\text{imp}}$ and $\pi^{\text{red}}$ are learnt with the two decomposed objective independently as in equation 9. When we combine these two learned policies together (as in Algorithm \ref{alg:the_alg}, and eqn.\ref{eqn:pobrl}), these two policies complement each other and formulate the optimal search strategy $\pi^{\text{PoBRL}}$.

\section{Details on the training and testing dataset}

We evaluate the performance of our proposed strategy on the following two datasets 

In the first experiment, we train and test on the Multi-News dataset \cite{Multinews}. This dataset contains 44972 training examples, and 5622 examples for testing and also 5622 examples for validation. Each summary is written by professional editor.  We follow the testing procedure as in \cite{Multinews}, and the dataset can be found in github link \footnote{https://github.com/Alex-Fabbri/Multi-News}.

In the second experiment, we test on the DUC-04 \cite{duc05} dataset. This dataset contains 50 clusters with each cluster having 10 documents. For each cluster, the dataset provide 4 human generated ground-truth summaries. Because it is a evaluation-only dataset, we follow the same procedure as in \cite{Multinews}\cite{lebanoff-etal-2018-adapting} to train our network with the CNN-DM \cite{CNNDM}, and then test on the DUC-04 dataset. The CNN-DM dataset contains 287,113 training, 13,368 validation, and 11,490 testing examples. It can be downloaded from here \footnote{https://github.com/abisee/cnn-dailymail}, and the DUC 04 dataset can be downloaded from here \footnote{https://duc.nist.gov/duc2004/}.

\section{Training details}
\textbf{Our Model}. 
For our hierarchical sentence extractor, we instantiate the temporal CNN module with a setup of 1-D single layer convolutional filters of the size of 3, 4, and 5, respectively. Each input sentence is first converted to a vector representation by a word2vec matrix with a output dimension of 128. Then, each sentence is encoded by concatenating the output of the each window from the temporal-CNN model.

Next, we instantiate the article-level encoder with a 2 layers bidirectional LSTM with 256 hidden units. We use this same configuration also for the encoder and decoder pointer-network as well. For the supervise learning (as the warm start), we trained the network with Adam optimizer with a learning rate of 1e-3 and with a batch size of 64. For the reinforcement learning part, we trained the network with actor-critic and with a learning rate of 1e-7,  $\gamma=0.99$ and a batch size of 32. 

The entire training time takes about 24.05 hours on a T4 GPU. All hyper-parameters are tuned with the validation set. For the batch size, we search over the values of [32, 128], and for the learning rate (supervised learning part), we search over the values of [1e-3, 1e-4]. For the learning rate in the reinforcement learning part, we search over the values of [1e-3, 1e-7]. For hyperparmeters searching, we ran 1 trial for one set of parameters.  After selecting the values for learning rates and batch size, finally, we tune $\lambda$. Here, we only tried $\lambda$ for the values of [0.8, 0.9], and we select the $\lambda$ value base on its performance on the validation set. For our $\lambda = \lambda^{\text{adv}}_t$, no tuning is required. 

Besides the learning rates and batch size, our overall model has 1 parameter (if using a fix value of $\lambda$) to tune. On the other hand, if using a adaptive $\lambda = \lambda^{\text{adv}}_t$, no tuning is required. 

The overall runtime (for summarizing the Multi-News dataset) is about 8.25 minutes (for $\lambda= 0.8, 0.9$), and 10.5 minutes for $\lambda = \lambda^{\text{adv}}_t$. For the DUC-04 dataset, the runtime is about 1.80 minutes (for $\lambda= 0.8, 0.9$), and 2.12 minutes for $\lambda = \lambda^{\text{adv}}_t$.

For the \textbf{baselines}, we used the exact setup as in their original papers.

\end{appendices}

\end{document}